# Methodology to Create Analysis-Naïve Holdout Records as well as Train and Test Records for Machine Learning Analyses in Healthcare

Michele Bennett, PhD; Mehdi Nekouei, PhD; Armand Prieditis, PhD; Rajesh Mehta, RPh;

Ewa Kleczyk, PhD; Karin Hayes

May 6, 2022

## ABSTRACT

It is common for researchers to holdout data from a study pool to be used for external validation as well as for future research, and the same desire is true to those using machine learning modeling research.  For this discussion, the purpose of the holdout sample it is preserve data for research studies that will be *analysis-naïve* and randomly selected from the full dataset. Analysis-naïve are records that are not used for testing or training machine learning (ML) models and records that do not participate in any aspect of the current machine learning study. The methodology suggested for creating holdouts is a modification of k-fold cross validation, which takes into account randomization and efficiently allows a three-way split (holdout, test and training) as part of the method without forcing. The paper also provides a working example using set of automated functions in Python and some scenarios for applicability in healthcare.

### Declaration

- Authors are affiliated with Symphony Health, a division of Icon, plc.
- Dr. Kleczyk is also an Affiliated Graduate Faculty in the School of Economics at the University of Maine, Orono, Maine
- Dr. Bennett is also an Adjunct Professor, within the Graduate and Doctoral Programs in Data Science, Computer Science, and Business Analytics at Grand Canyon University
- Competing Interest: The authors declare that they have no competing interests.
- Funding: Authors work for Symphony Health, ICON plc Organization.
- Author contact: Michele Bennett; mbennett1107@gmail.com

### Need for Sample Holdouts

It is common for researchers to hold out data from the study pool to be used for external validation as well as for future research, and the same desire is true to those using machine learning modeling as part of their patient research (Schorfheide & Wolpin, 2013). For this discussion, the purpose of the holdout sample it is preserve a subset of patients for future research studies that will be *analysis-naïve* and randomly selected from a group of patients diagnosed with a particular disease in an open medical and hospital claims dataset from Symphony Health (Symphony Health, 2022). Analysis-naïve patients are those records not used for testing or training machine learning (ML) models and means that these patient record do not participate in any aspect of the current machine learning study. This definition is slightly different than holdout records where after the model is trained (training data) and the model is used on the holdout records to test the model on unseen data (test data) (Cho et al. (2016). This *analysis-naïve* group will be called *holdouts* to conform with standard ML language (Martins, 1989; Scheider, 1997).

### Why Use a Modification of K-Fold Cross Validation

The methodology suggested for creating holdouts is a modification of k-fold cross validation, which is a machine learning standard method for splitting records into test and train and ensures that every record has the opportunity to be in each category as per randomization (Martins, 1989; Scheider, 1997). The methodology was chosen because it takes into account randomization (Scheider, 1997), and while it is a performance evaluating technique as well, it efficiently separates records to allows the use of the folds for the required holdouts as well as test and train groups while simultaneously offering the ability to determine the optimized size of the groups needed for a study and sample creation evaluation (Dietterich, 1998; Raschka, 2018;

Wong, 2015). In addition, it allows a three-way split as part of the method, without forcing.

While this approach is  used more for neural network and certain regression designs, it can be taken advantage of for any machine learning design and study (Wiens et al., 2008). To provide a working example, the methodology will be explained using set of automated functions in Python for each of execution of the cross validation which also visualization of the output as part of the process (scikit-learn, 2021), which also is helpful in the configuration process.

## Creating Train, Test, and Hold-Out Records

The standard k-fold cross validation will be extended (see Figure 1) as the data

partitioning strategy in order to create three groups - holdout for future use, test, and train groups

(James et al., 2017) rather than the typical two groups. Figure 1 was adapted from the original in the Python documentation (scikit-learn, 2021) to demonstrate and include the allocation of a fold to hold out data.

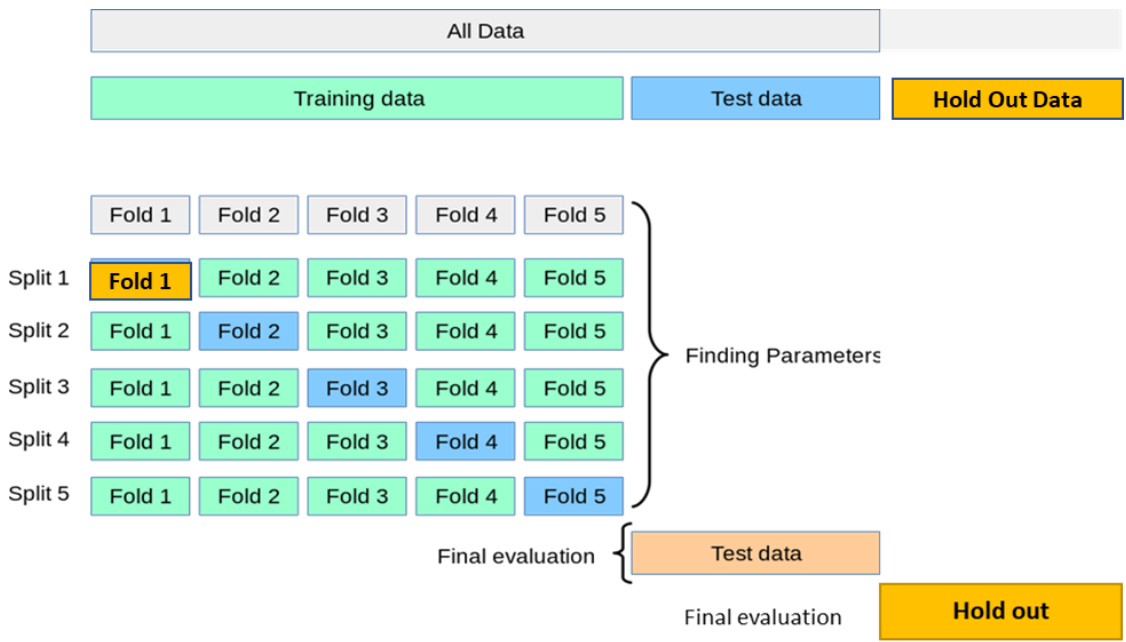

Figure 1. Illustration of k-fold cross validation with extension for hold outs (scikit-learn, 2021)

The k-fold cross validation process described below is commonly used across a wide array of machine learning projects. This approach involves randomly dividing patient records into k groups, or folds, of approximately equal size. The first fold is treated as a holdout set, the second is the test set, and the training set is created on the remaining k − 2 folds (Brownlee, 2018; Efron & Tibshirani, 1997; Martins, 1989; Scheider 1997).

**Process for Running K-Fold**

- Shuffle  (which is literal and therefore, using the *shuffle* function in Python) the data to create a randomized version of the data, and in this case, the entire dataset will be randomized (scikit-learn, 2021)

- Split the data into k groups (scikit-learn, 2021)

- For each unique group:

  - Label the group as a holdout or test records. This is a critical step to ensure that the holdout records are kept separately and as analytics-naïve records. Fold 1 in Figure 1 is the holdout group.

  - For example, the records could be labeled by adding a field in the database that will contain a hash key, identifying the disposition of that records (holdout, train, or test) so that records cannot become confused.

  - While holdouts are selected and labeled once, test and train records could be labeled differently for subsequent model iterations using the same data, as it is possible that a record can be test in one study or project and train in another, but the holdout of analytics-naïve records would remain consistent.

  - The remaining groups are considered the training data set, and each record  is labeled as train using the same hash key approach shown in Figure 1.

- Modeling is then run on the training records and evaluated on the test record and AutoML can be used to fit a one-time model to balance or equate the size of the folds but not to inform or otherwise bias modeling or analysis. Therefore, individual variables would not be to evaluate the features of this modeling activity at this time (Rodriguez et al., 2009). The evaluation skill score (*F1* score - overall correctness the model based on precision and recall) are kept and the model is discarded.

- This model is used to create the folds and is not used to evaluate objectives or for analysis except to balance skill scores, so it is discarded after the dataset is labeled. However, the skill score and results error terms will be also used as part of the sample frame to determine sample sizes needed.

- The model evaluation skill scores are kept for use in the Sample Frame (see below).

- The result is a split and labeled dataset with holdout, test and train records and a corresponding hash key in the newly added field for each record in the dataset.

**Configuration of K**

The configuration of k is important to ensure representativeness and three common tactics for choosing a value for k can be used with this overall methodology, They include:

- *Representative:* The value for k is chosen such that each train/test group of data samples statistically representative. A paired t-test, specifically a 5x2cv paired t-test, is used where the mean, variance and error terms are compared of 50%-50% test/train (or 2-fold) tests over 5 times selecting the k with lowest errors  (Dietterich, 1998).

- *k=10*: The value for k is fixed to 10, a value that has been found through experimentation to generally result in a model skill estimate with low bias a modest variance, which is typical when there are a large number of variables and a known range of values from which to approximate variance  This is a common choice as studies have found that is provide good trade-off of low computational cost and low bias in an estimate of model performance (Raschka, 2018) and might be valuable for studies and projects where more is known about features and variables a priori.

- $k=n$: The value for k is fixed to n, where n is the size of the dataset (excluding the hold-out subset) to give each observation an opportunity to be used as test set. This approach is called leave-one-out cross-validation (LOOCV), which computationally expensive, but can provide a good estimate of model performance given the available data (Wong, 2015). This can be useful for smaller datasets or where some data is used for holdouts, and analysis is not using all available data.

To balance and minimize bias, error rate, and data variability, the .632+ bootstrap method (Efron & Tibshirani, 1997) with the mean of the model skill scores is a preferred method to select the appropriate k to ensure k groups with the same number of samples, such that the sample of model skill scores are all equivalent (James et al., 2017).

### Putting K-Fold Cross Validation into Practice for Healthcare

This approach may be useful in healthcare where the underlying data, market conditions, or some aspect of the brand or promotion may shift in the future, and both reevaluation of the model as well as for new research present valid use cases for the holdout sample group (Tougui et al., 2021).  Some specific scenarios include the following:

- New data is integrated or added to existing data set thus the model should be re-tested (Schorfheide & Wolpin, 2013).

- If the market condition surrounding the original modeling have changed, reevaluation is warranted and the holdout may be useful (Saeb et al., 2021).  For example, if a competitive brand is newly launched or has a new indication; COVID-19 changed how physicians are treating a condition or how patients may receive care (i.e. telemedicine may be different that in office visits); patients were late to receive screening or treatment and that has long term impact on disease progression and future treatment (Saeb et al., 2021; Tohka & van Gils, 2021); or large payer with a restrictive formulary changes preferred products in a particular region and patients will be lost to health plan changes over time, thus decreasing the cohort (Delen et al., 2009).

- Promotion or product of interest changes over time whether there are new campaigns, new indications or brands. For example, the brand itself may change the model needs if it expands the market or changes the treatment journey for many patients. Omni-channel campaigns can drive awareness and increase the number of patients who require treatment (Tohka & van Gils, 2021; Tougui et al., 2021).

- A patient's journey from diagnosis through treatments and disease progression changes over time, and the original holdout sample will still need to be relevant for the question of interest at the time they are used (Mihalik et al, 2020; Schorfheide & Wolpin, 2013). For example, if a model is used to predict first line initiation and the cohort of diagnosed untreated patients may have progressed through later lines, the holdout patient group will still need to be valid for model evaluation (Tohka & van Gils, 2021).

### For Consideration

In addition, record that are to be designated for test an earlier study may also be considered analysis-naïve patients, as they were not used in the model training process and machine learning methodologists do consider test data as holdouts (Bauer & Kohavi, 1999;

Mihalik et al., 2020; Raschka, 2018).  For smaller datasets, these records may be valuable to follow up studies while still remaining analysis-naïve data (Wong, 2015).

## Conclusion

For those studies that wish to maintain a cohort of records that remain analysis-naïve and available for either future research or external validation, a modified k-fold cross validation provides a best-practice based approach that is straightforward to execute in Python or other machine learning toolsets (scikit-learn, 2021).

**References**

Bauer, E. & Kohavi, R. (1999). An empirical comparison of voting classification algorithms:

Bagging, boosting, and variants, *Machine Learning, 36*, 105-139. doi:

10.1023/A:1007515423169

Brownlee , J. (2018). A gentle introduction to k-fold cross-validation. Retrieved from

https://machinelearningmastery.com/k-fold-cross-validation/

Bzdok, D., Krzywinski, M., & Altman, N. (2017). Machine learning: a primer. *Nature methods,*

*14*(12), 1119–1120. doi:10.1038/nmeth.4526

Cho et al. (2016). How much data is needed to train a medical imaging deep learning system to

achieve necessary high accuracy? *ArXiv*. https://arxiv.org/abs/1511.06348v2

Delen et al. (2009). Analysis of healthcare coverage: A data mining approach. *Expert Systems*

*with Applications, 36*(2), 995-1003. doi: 10.1016/j.eswa.2007.10.041

Dietterich, T. G. (1998). Approximate statistical tests for comparing supervised classification

learning algorithms. *Neural Computation 10*(7), 1895-1923. doi:

10.1162/089976698300017197

Efron B., & Tibshirani, R. (1997). Improvements on cross-validation: The .632+ bootstrap

method. *Journal of the American Statistical Association, 92* (438), 548-560. doi:

10.2307/2965703

Figueroa, R. L., Zeng-Treitler, Q., Kandula, S. & Ngo, L. H. (2012). Predicting sample size

required for classification performance. *BMC Med Inform Decis Mak 12* (8).

https://doi.org/10.1186/1472-6947-12-8

James et al., (2017). *An Introduction to Statistical Learning: with Applications in R (Springer*

*Texts in Statistics).*  Springer: New York.

Li et al. (2019). Using virtual samples to improve learning performance for small datasets with

    multimodal distributions. *Soft Computing 23*, 11883 – 11900. doi:

    10.1016/j.ress.2021.108114

Mihalik et al. (2020). Multiple holdouts with stability: Improving the generalizability of

    machine learning analyses of brain–behavior relationships. *Biological Psychiatry, 87*(4),

    368-376. doi: 10.1016/j.biopsych.2019.12.001

Martins. P. (1989). Resampling methods. Retrieved from https://www.paul-

    martins.com/posts/islr/04_resampling/resampling.pdf

Poczas, B. (2015). Advance introduction to machine learning. Retrieved from

    http://www.cs.cmu.edu/~bapoczos/Classes/ML10715_2015Fall/slides/VCdimension.pdf

Raschka, S. (2018). Model evaluation, model selection, and algorithm selection in machine

    learning. *Cornell University Press*. 1-49. Retrieved from

    https://arxiv.org/pdf/1811.12808.pdf

Rodriguez et al. (2009). Sensitivity analysis of k-Fold Cross Validation in Prediction Error

    Estimation. *IEEE Transactions on Pattern Analysis and Machine Intelligence, 32*(3),569-

    575. doi: 10.1109/TPAMI.2009.187

Saeb et al. (2017). The need to approximate the use-case in clinical machine learning.

    *Gigascience, 6*(5), 1-9. doi: 10.1093/gigascience/gix019

Scheider, J. (1997). Cross Validation. Retrieved from

    https://www.cs.cmu.edu/~schneide/tut5/node42.html

Schorfheide, F., & Wolpin, K., I., (2013). To hold out or not to hold out. *Research in Economics,*

    *70*(2), 332-345. doi: 1016/j.rie.2016.02.001

Scikit.learn (2021). Cross validation: evaluating and estimating performance. https://scikit-learn.org/stable/modules/cross_validation.html#cross-validation-evaluating-estimator-performance

Sontag. E. D. (1998). VC dimension of neural networks. Retrieved from http://www.vision.jhu.edu/teaching/learning/deeplearning18/assets/Sontag-98.pdf

Symphony Health (2022). Integrated DataVerse (IDV). Retrieved from https://symphonyhealth.com/insights/idv-overview

Tohka, J., & van Gils, M. (2021). Evaluation of machine learning algorithms for health and wellness applications: A tutorial. *Computers in Biology and Medicine, 132*, 104324. doi: 10.1016/j.compbiomed.2021.104324

Tougui, I., Jilbab, A., & Mhamdi, J. E. (2021). Impact of the choice of cross-validation techniques on the results of machine learning-based diagnostic applications. *Healthcare Informatics Research, 27*(3), 189–199. doi: 10.4258/hir.2021.27.3.189

Viering, T., & Loog, M. (2021). The shape of learning curves: A review. *ArXiv, abs/2103.10948.*

Wiens et al. (2008). Three way k-fold cross-validation of resource selection functions. *Ecological Modeling, 212*(3-4), 244-255. doi:10.1016/j.ecolmodel.2007.10.005

Wong, T-T. (2015). Performance evaluation of classification algorithms by k-fold and leave-one-out cross validation. *Pattern Recognition, 48*(9), 2839-2846. doi: 10.1016/j.patcog.2015.03.009

Zhang, Y., & Ling, C. (2018). A strategy to apply machine learning to small datasets in materials science. *Nature: NPJ Computational Materials, 25*, https://doi.org/10.1038/s41524-018-0081-z